\documentclass[10pt]{article}

\usepackage[margin=1in]{geometry}
\usepackage[numbers,sort&compress]{natbib}

\usepackage{iftex}
\ifPDFTeX
  \usepackage[utf8]{inputenc}
  \usepackage[T1]{fontenc}
\fi

\usepackage{mathtools}
\usepackage{amsmath}
\usepackage{amsfonts}
\usepackage{amssymb}
\usepackage{bm}
\usepackage{physics}

\usepackage{booktabs}
\usepackage{multirow}
\usepackage{siunitx}
\usepackage{graphicx}

\usepackage{microtype}
\usepackage{parskip}

\usepackage{hyperref}
\usepackage{url}
\usepackage[nameinlink,capitalize]{cleveref}

\newcommand{\myparagraph}[1]{\par\vspace{1mm}\noindent\textbf{#1}}

\title{Guided Diffusion Sampling for\\ Precipitation Forecast Interventions}
\date{}

\author{%
  Ayumu Ueyama$^{1}$,\quad Kazuhiko Kawamoto$^{1}$,\quad Hiroshi Kera$^{1,2}$\thanks{Corresponding author: Hiroshi Kera (\texttt{kera@chiba-u.jp}).}\\[0.6em]
  $^{1}$ Chiba University \qquad $^{2}$ National Institute of Informatics\\[0.8em]
  Code: \url{https://github.com/kawakera-lab/guided-diffusion-weather-intervention}
}

\begin{document}

\maketitle

\begin{abstract}
    Extreme precipitation causes severe societal and economic damage, and weather control has long been discussed as a potential mitigation strategy.
    However, to the best of our knowledge, perturbation-based interventions for weather control using data-driven weather forecasting models have not yet been explored.
    While adversarial attacks also generate perturbations that alter forecasts, they aim to exploit model artifacts and do not account for physical plausibility.
    In this paper, we propose a gradient-based guidance framework for precipitation-reduction interventions through diffusion sampling in diffusion-based weather forecasting models.
    Instead of directly perturbing atmospheric states, our method steers the diffusion sampling trajectory, enabling precipitation reduction while maintaining consistency with the atmospheric distribution.
    To assess physical plausibility, we evaluate from three perspectives: (i) vertical and variable-wise perturbation profiles, (ii) latent-space trajectory deviation, and (iii) cross-model transferability.
    Experiments on extreme precipitation events from WeatherBench2 demonstrate that our method achieves effective precipitation reduction while yielding more physically plausible interventions than adversarial perturbations.
\end{abstract}

\section{Introduction}
Extreme precipitation has significant impacts on society, the environment, and the economy~\cite{fischer2016observed, papalexiou2019global}. 
Although challenging, several attempts have been made to mitigate these effects, such as operational field experiments (e.g., Project Stormfury~\cite{willoughby1985project} for hurricane suppression) and theoretical approaches using numerical weather prediction (NWP) models~\cite{hiraga2025numerical, henderson20054d, YuehuaPeng2023WeakConstraint}.
Hoffman’s framework characterizes weather intervention as a feedback control problem, in which small, carefully designed perturbations steer atmospheric evolution toward desired outcomes~\cite{hoffman2002controlling}.
However, NWP-based approaches are computationally expensive, thereby limiting real-time exploration of intervention strategies.

Recent advances in data-driven weather forecasting models provide a promising framework for intervention.  
Unlike time-consuming NWP models, data-driven models can generate forecasts in near real-time while achieving competitive forecast accuracy~\cite{Bi2023Accurate, Lam2023Learning, Bonev2023Spherical, Kochkov2024Neural, Nguyen2024Scaling, Price2025Probabilistic, Bodnar2025Foundation}. 
This enables efficient design and evaluation of interventions. 
Specifically, a small perturbation to the current atmospheric state can be iteratively computed and refined to achieve the desired forecast change, such as reducing precipitation in a target region. 
However, to the best of our knowledge, no prior work has investigated weather intervention using data-driven weather forecasting models. 
The most closely related work is AOWF~\cite{Imgrund2025Adversarial}, which addresses adversarial attacks~\cite{Goodfellow2014Explaining, Madry2018Towards, Dong2018Boosting} on forecasting model outputs, showing the potential security risks of forecasting models. 
However, the designed perturbations exploit model artifacts that induce unrealistic behavior and are not physically plausible, as will be shown in our experiments (see~\cref{sec:structural_properties_of_the_interventions,sec:latent_space_deviation_under_intervention,sec:intervention_transferability}).

In this study, we propose the first comprehensive framework for precipitation-reduction intervention with diffusion-based weather forecasting models. 
In particular, we introduce gradient-based guidance to steer the diffusion model's sampling trajectory and generate interventions that are more physically plausible than adversarial perturbations.
For evaluation, we construct a new dataset of extreme precipitation events from WeatherBench2~\cite{Rasp2024WeatherBench}, and assess both the precipitation reduction and the physical plausibility of the generated interventions. 
The physical plausibility is investigated from three aspects: (i) the profiles of the interventions across atmospheric variables and pressure levels, (ii) the deviation in the latent space of the diffusion model under intervention, and (iii) the transferability of the intervention to other data-driven models.

\begin{itemize}
 \item We introduce the first comprehensive study of precipitation-reduction interventions using data-driven models. We construct a new dataset of extreme precipitation events from WeatherBench2~\cite{Rasp2024WeatherBench}, propose a gradient-based guidance framework for interventions with diffusion-based weather forecasting models, and introduce several evaluation metrics to quantify the physical plausibility of the generated interventions.
 \item Our method generates perturbations by guiding the sampling trajectory of the diffusion model, rather than directly perturbing atmospheric states as in adversarial attacks. This approach achieves distributional consistency in latent space, leading to more physically plausible interventions.
 \item Experiments show that the proposed method achieves competitive performance in precipitation reduction compared to adversarial perturbations. Importantly, our analysis supports the physical plausibility of the proposed method, whereas adversarial perturbations lead to significant divergence in the diffusion model's latent space and to unreasonable perturbation profiles.
\end{itemize}

\section{Related Work}
\subsection{Weather Control and Weather Modification}
Weather control and weather modification aim to study whether to mitigate extreme events~\cite{Bruintjes1999AReview, abe2025historical}.
Prior studies have examined intervention strategies for typhoons and extreme precipitation through both operational experiments and conceptual analyses~\cite{alamaro2006preliminary, klima2012hurricane, hiraga2025numerical}.
Project Stormfury attempted to weaken hurricanes by seeding them with silver iodide from aircraft~\cite{willoughby1985project}.
Recently, Japan launched the Moonshot Research and Development Program to explore technological and societal pathways to reduce heavy rainfall intensity by 2050.
Theoretical studies have explored mitigation of extreme precipitation and typhoons via perturbation design and numerical experiments using weather forecasting models~\cite{henderson20054d, YuehuaPeng2023WeakConstraint}.
Hoffman~\cite{hoffman2002controlling} characterized the atmosphere as a chaotic dynamical system and suggested that small perturbations may influence forecast evolution.
However, such approaches rely on NWP and incur computational costs, limiting large-scale, rapid experimentation.
To overcome this limitation, we investigate intervention through input perturbation in data-driven weather forecasting models.

\subsection{Deep Learning for Weather Forecasting}
Data-driven weather forecasting models trained on reanalysis data~\cite{Hersbach2020ERA5} have advanced as alternatives to NWP~\cite{bauer2015quiet}.
Pangu-Weather~\cite{Bi2023Accurate} introduced a Vision Transformer~\cite{Dosovitskiy2021An} variant tailored to the Earth’s three-dimensional structure, becoming the first to surpass NWP performance.
This result catalyzed the emergence of practical data-driven models, including GraphCast~\cite{Lam2023Learning} and GenCast~\cite{Price2025Probabilistic}, which are establishing a new paradigm in weather forecasting~\cite{Bonev2023Spherical, Kochkov2024Neural, Nguyen2024Scaling, Bodnar2025Foundation}. 
These models reduce inference time and computational cost compared to NWP models, enabling real-time forecasting. 
GenCast~\cite{Price2025Probabilistic} is a diffusion-based conditional generative model~\cite{Sohl-Dickstein2015Deep, Song2021Score-Based, Karras2022Elucidating} 
that enables sampling from the data distribution of atmospheric states given past observations. 
Accordingly, we adopt GenCast as the forecasting backbone for generating physically plausible perturbations.

\subsection{Adversarial Attacks in Weather Forecasting}
Adversarial attacks manipulate model outputs by injecting small perturbations into input data~\cite{Goodfellow2014Explaining, Madry2018Towards, Dong2018Boosting}.
In weather forecasting, observational noise and data assimilation uncertainty can mimic natural measurement errors~\cite{Imgrund2025Adversarial}.
This property makes weather forecasting models susceptible to adversarial attacks via small input perturbations~\cite{Deng2025Fable, Arif2025ForecastingFails, Deng2025Adversarial}.
Imgrund et al.~\cite{Imgrund2025Adversarial} showed that perturbations equivalent to tampering with a single weather satellite can misdirect forecasts.
Such vulnerabilities pose serious security risks and may undermine trust in operational weather forecasting systems.
In contrast, this study seeks perturbations that reflect physically plausible atmospheric state variations, rather than adversarial perturbations that exploit model artifacts.

\section{Problem Setting}\label{sec:problem_setting}
This section presents the problem setting for weather intervention with data-driven weather forecasting models and the associated challenges. 

\subsection{Problem Formulation}
Let $\mathbf{X}^t \in \mathbb{R}^{H \times W \times V}$ be the atmospheric state at time $t$, containing values of $V$ variables at each of the $H \times W$ grid points.
A one-step forecasting model $f : \mathbf{X}^{t} \mapsto \hat{\mathbf{X}}^{t+1}$ outputs $\hat{\mathbf{X}}^{t+1}$, the atmospheric state at time $t+1$.
\footnote{Data-driven forecasting models can use multiple time steps for input. For example, GenCast~\cite{Price2025Probabilistic} uses the current and previous states, i.e., $f_{\mathrm{GenCast}} : (\mathbf{X}^{t-1}, \mathbf{X}^{t}) \mapsto \hat{\mathbf{X}}^{t+1}$. Our formulation can be readily extended to this case, but for simplicity of notation, we assume that forecasting models use only the current state.} 
Recursive application of $f$ yields a $j$-step forecast $\hat{\mathbf{X}}^{t+j}$ for $j > 1$.

The intervention task aims to reduce precipitation in the target region $R$ at the target time step $t+T$ and can be formulated as follows. 
\begin{align} \label{eq:problem_formulation}
    &\min_{\bm{\delta}^{t+1}}\ \mathcal{L}\qty( \mathcal{P}_{R}\qty[\hat{\mathbf{X}}^{t+T}] ), \\
    &\mathrm{where} \quad \hat{\mathbf{X}}^{t+T} = \underbrace{(f\circ \cdots \circ f)}_{T-2 \text{ times}}\qty(f(\hat{\mathbf{X}}^{t+1} + \bm{\delta}^{t+1})), \label{eq:problem_formulation_dyn}
\end{align}
where $\mathcal{L}$ is a loss function and $\mathcal{P}_{R}$ restricts the field to region $R$.

It is worth noting that the perturbation is added only at the time step $t+1$. 
This reflects a feasibility requirement: the intervention cannot be applied immediately and requires a setup time based on the forecasting (e.g., $\hat{\mathbf{X}}^{t+1}$). 
Moreover, while the intervention can be applied only briefly, its effect must persist to influence the forecast at $t+T$.
In our formulation, the intervention $\bm{\delta}^{t+1}$ is applied at time $t+1$ in $f(\hat{\mathbf{X}}^{t+1} + \bm{\delta}^{t+1})$.

Note that the adversarial attack on the forecasting model in prior work~\cite{Imgrund2025Adversarial} assumes poisoning the input, and thus the attack can be applied immediately to any region. 
Any time delay is not considered.

\subsection{Physical Plausibility of Intervention}
One possible approach to computing the perturbation $\bm{\delta}^{t+1}$ in \cref{eq:problem_formulation} is gradient-based optimization as adversarial attacks (e.g., PGD~\cite{Madry2018Towards} and AOWF~\cite{Imgrund2025Adversarial}). Namely, we start with an initial perturbation $\bm{\delta}^{t+1}_0$ and update it iteratively by gradient descent as follows: for $i=0,1,2,\ldots$,
\begin{align} \label{eq:direct_state_space_perturbation}
    \bm{\delta}^{t+1}_{i+1}
    \leftarrow \bm{\delta}^{t+1}_i - \eta\cdot \mathrm{Clip}\qty[\nabla_{\bm{\delta}^{t+1}_i} \mathcal{L}\qty( \mathcal{P}_{R}\qty[\hat{\mathbf{X}}^{t+T}] )],
\end{align}
where $\eta > 0$ is a step size, and $\mathrm{Clip}[\,\cdot\,]$ is a clipping operator to ensure the perturbation stays within a prescribed bound.

This approach directly optimizes the perturbation $\bm{\delta}^{t+1}$ to minimize the loss function $\mathcal{L}\qty( \mathcal{P}_{R}\qty[\hat{\mathbf{X}}^{t+T}] )$ by gradient descent. 
While this may generate perturbations that can alter the forecast, it does not guarantee the physical plausibility of the intervention. 
Indeed, our experiments show that such perturbations are meteorologically unrealistic in terms of their perturbation profiles and distributional consistency (see~\cref{fig:quantitative_intervention,fig:latent_evaluation,fig:latent_PCA_evaluation}). 
The direct optimization of the perturbation is too aggressive, and the perturbed forecast can immediately diverge from the latent distribution of the atmospheric state. 

In the next section, we introduce an indirect approach to generate perturbations by guiding the sampling trajectory of a diffusion-based forecasting model. 
The diffusion process generates the atmospheric state from noise, using the atmospheric distribution approximated by the model. 
Guiding this process helps the perturbed forecast remain closer to the atmospheric distribution.

\section{Method}\label{sec:method}
This section describes weather intervention via diffusion sampling with gradient-based guidance under the atmospheric distribution approximated by GenCast~\cite{Price2025Probabilistic}. 
Instead of perturbing forecast states directly, the proposed method modifies intermediate denoising states during diffusion sampling to stay closer to the data manifold~\cite{Dhariwal2021Diffusion, Ho2022Classifier, Jeanneret2022Diffusion, Weng2024Fast}.

In the following, we first summarize diffusion sampling in GenCast and then introduce the proposed gradient-based guidance mechanism.

\subsection{Diffusion Forecasting in GenCast}

The GenCast model is a diffusion-based forecasting model that captures the conditional distribution
$
p(\mathbf{X}^{t+1} \mid \mathbf{X}^{t-1}, \mathbf{X}^{t})
$
through diffusion sampling.
Rather than forecasting the next state $\hat{\mathbf{X}}^{t+1}$ directly, it generates a normalized residual
$
\mathbf{Z}^{t+1}
$
relative to the current state $\mathbf{X}^t$.
The forecast is reconstructed as
\begin{align}\label{eq:forecast}
\hat{\mathbf{X}}^{t+1}
=
\mathbf{X}^t + \mathbf{S}\mathbf{Z}^{t+1},
\end{align}
where $\mathbf{S}$ is the inverse normalization matrix precomputed from training data.

The diffusion sampling process proceeds over noise levels
$\{\sigma_i\}_{i=0}^{N}$.
Starting from an initial noisy residual
$\mathbf{Z}^{t+1}_0$,
it is iteratively refined for $i = 0, 1, \ldots, N-1$.
At each step $i$, a neural network denoiser $d$ estimates a denoised residual $\hat{\mathbf{Z}}^{t+1}_i$ from the $i$-th step noisy residual $\mathbf{Z}^{t+1}_i$:
\begin{align}\label{eq:diffusion_sampling}
\hat{\mathbf{Z}}^{t+1}_i
=
d(\mathbf{Z}^{t+1}_i; \mathbf{X}^{t-1}, \mathbf{X}^{t}, \sigma_i).
\end{align}
This estimate is then used in a deterministic update rule to obtain the next noisy residual
$\mathbf{Z}^{t+1}_{i+1}$.
After $N$ refinement steps, the final residual
$\mathbf{Z}^{t+1} := \mathbf{Z}^{t+1}_N$
is obtained, and the forecast is reconstructed by~\cref{eq:forecast}.

\subsection{Diffusion Sampling with Gradient-Based Guidance}
\begin{figure}[tb]
  \centering
  \includegraphics[width=0.8\linewidth]{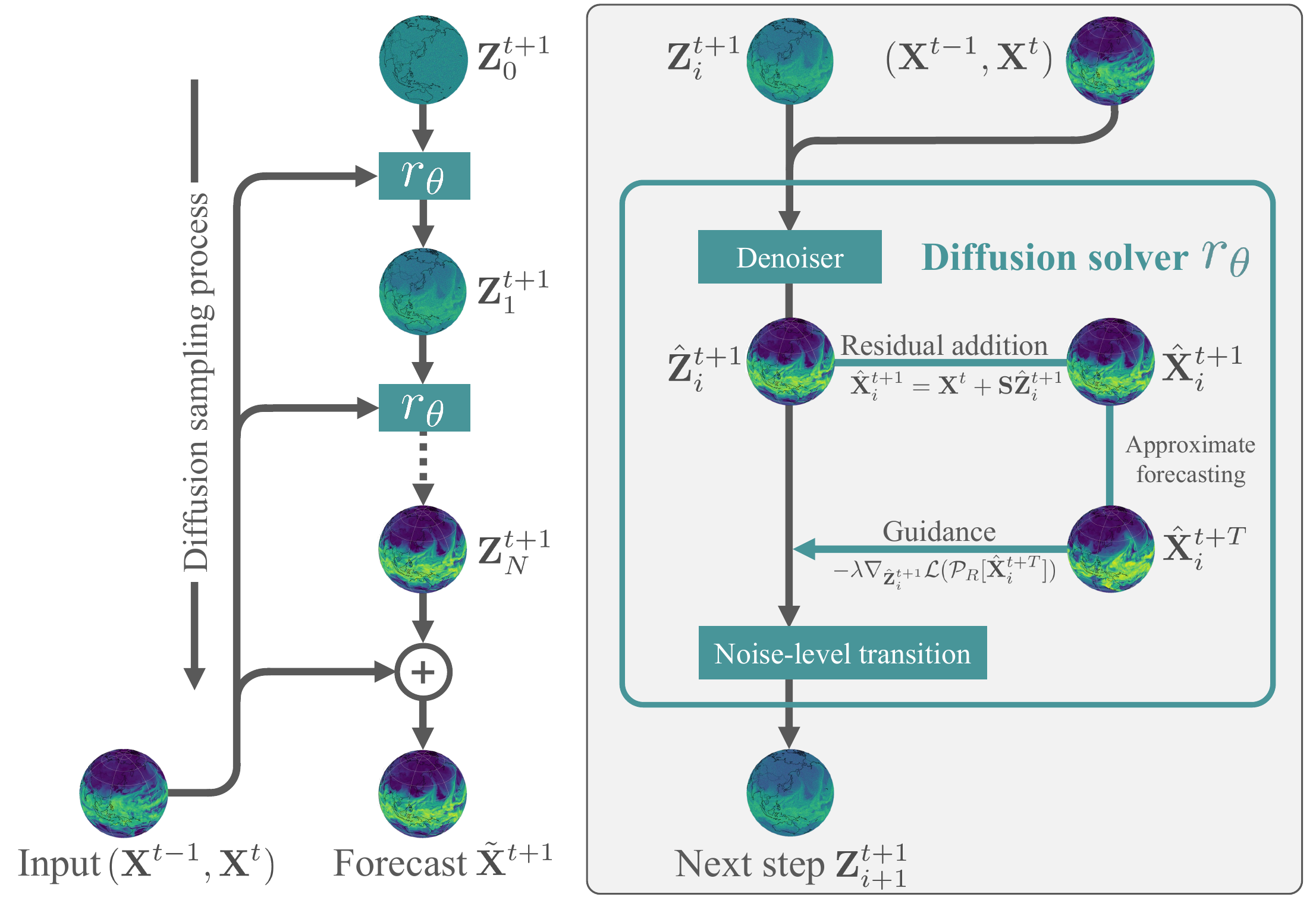}
  \caption{
    Overview of the proposed method.
    In GenCast diffusion sampling, the denoised residual $\hat{\mathbf{Z}}^{t+1}_i$ is adjusted by the gradient of a control objective,
    yielding a guided residual $\tilde{\mathbf{Z}}^{t+1}_i$.
    The final guided forecast $\tilde{\mathbf{X}}^{t+1}$ is reconstructed from the modified residual.
  }
  \label{fig:method}
\end{figure}
The proposed method guides the diffusion sampling process of the GenCast model, as in~\cref{eq:diffusion_sampling}, toward the desired atmospheric state, as shown in~\cref{fig:method}. Specifically, with the loss $\mathcal{L}$ in~\cref{eq:problem_formulation}, we perturb the $i$-th denoised residual estimate $\hat{\mathbf{Z}}^{t+1}_i$ as
\begin{align}\label{eq:gradient_guided_diffusion_sampling}
  &\tilde{\mathbf{Z}}^{t+1}_i = \hat{\mathbf{Z}}^{t+1}_i - \lambda \nabla_{\hat{\mathbf{Z}}^{t+1}_i}\mathcal{L}\qty(\mathcal{P}_{R}\qty[\hat{\mathbf{X}}^{t+T}_i]), \quad (i = 0, 1, \ldots, N-1),
\end{align}
where $\hat{\mathbf{X}}^{t+T}_i$ denotes the forecast obtained by propagating the denoised residual $\hat{\mathbf{Z}}^{t+1}_i$. 
The hyperparameter $\lambda > 0$ is a guidance scale. To avoid excessive perturbation, we omit the guidance at the final noise level $\sigma_N$.

The final perturbation $\bm{\delta}^{t+1}$ is the difference between the atmospheric states at time $t+1$ with and without guidance: 
\begin{align}
  \bm{\delta}^{t+1} = \tilde{\mathbf{X}}^{t+1} - \hat{\mathbf{X}}^{t+1}.
\end{align}

\myparagraph{Approximate inference.}
The computation of the perturbation $\bm{\delta}^{t+1}$ requires full backpropagation through the sampling steps and forecast rollout up to $t+T$, which is computationally expensive. To mitigate this, we introduce an approximate inference technique from prior work~\cite{Imgrund2025Adversarial}. 
Specifically, instead of handling all noise levels $\{\sigma_i\}_{i=0}^{N}$, we use only a few representative noise levels to compute the perturbation $\bm{\delta}^{t+1}$. 
The ordered noise levels $\{\sigma_i\}_{i=0}^{N}$ are partitioned into $n$ consecutive subsets, and one representative noise level is sampled from each subset.
This defines a one-step transition $\hat{\mathbf{X}}^{t+2}_i = \hat{f}(\mathbf{X}^{t}, \hat{\mathbf{X}}^{t+1}_i)$, where $\hat{f}$ is an approximate inference using the representative noise levels.
By replacing $f$ with $\hat{f}$ in~\cref{eq:problem_formulation_dyn}, we obtain the approximate future forecast at time $t+T$ (corresponding to $\hat{\mathbf{X}}^{t+T}_i$), from which the gradient can be efficiently computed.

\myparagraph{Computational complexity.}
We compare the cost of our method and AOWF~\cite{Imgrund2025Adversarial} in denoiser-pass counts per event.
AOWF runs $K$ PGD iterations, each performing $n$-step approximate inference in both forward and backward directions, totaling $Kn$ forward and $Kn$ backward passes.
Our method runs $N$ diffusion sampling steps; each step issues one denoiser call for sampling and, when guided, $n$-step approximate inference in both forward and backward directions, totaling $N + nN$ forward and $nN$ backward passes.
With $K{=}50$, $n{=}2$, and $N{=}20$~\cite{Price2025Probabilistic}, AOWF requires $100$ forward and $100$ backward passes whereas our method requires $60$ forward and $40$ backward passes, making our method theoretically about $2\times$ cheaper than AOWF.
This is consistent with the measured wall-clock time, $14.1$ seconds per event for AOWF versus $8.4$ seconds for ours.

\myparagraph{Loss function.}
We focus on precipitation reduction by minimizing the average precipitation within the target region.
The loss in~\cref{eq:problem_formulation} is defined as
\begin{align}\label{eq:loss_function}
  \mathcal{L}\qty(\mathcal{P}_{R}\qty[\hat{\mathbf{X}}^{t+T}]) = \frac{1}{|R|}
  \sum_{(p,q) \in R}
  \hat{\mathbf{P}}^{t+T}_{p,q}.
\end{align}
Here, $R$ denotes the set of latitude–longitude grid points within the target region, and $\hat{\mathbf{P}}^{t+T}$ denotes the forecasted precipitation field in $\hat{\mathbf{X}}^{t+T}$.

\section{Effectiveness of Interventions}
This section evaluates the effectiveness of the proposed method for precipitation reduction. 
Since no ground-truth observations exist after intervention, forecast validity cannot be assessed by forecast errors.
Therefore, the plausibility of the generated interventions is evaluated in \cref{sec:structural_properties_of_the_interventions,sec:latent_space_deviation_under_intervention,sec:intervention_transferability}.

\subsection{Setup}
We describe the experimental setup.
Unless otherwise specified, the same configuration is used in \cref{sec:structural_properties_of_the_interventions,sec:latent_space_deviation_under_intervention,sec:intervention_transferability}.

\myparagraph{Dataset.}
We collected extreme precipitation events from WeatherBench2~\cite{Rasp2024WeatherBench}.
We use WeatherBench2 data from 2022 at $1^\circ$ resolution, following the evaluation setup of AOWF~\cite{Imgrund2025Adversarial}, and select grid points whose precipitation anomaly relative to climatology exceeds $\tau = 0.06668$.
Here, $\tau$ is the spatial average of the 99th percentile of yearly maximum anomalies.
This threshold yields 219 extreme precipitation events; this is not a subsample but the entire set of events worldwide that satisfy the threshold in 2022, reflecting the intrinsic rarity of such events.
The 219 events nonetheless cover a wide range of locations such as Japan, Brazil, and India.
The target region is the $\pm 2^\circ$ neighborhood of each exceedance grid point in latitude and longitude.
Additional details are provided in the supplementary material.

\myparagraph{Model.}
We use GenCast~\cite{Price2025Probabilistic} as the forecasting model, which produces 12-hour forecasts in a single step.
Our evaluation focuses on the forecasted 12-hour total precipitation.
Perturbations are applied only to temperature and the u and v components of wind among the atmospheric input variables at time $t+1$.
We exclude specific humidity from the perturbed variables for feasibility reasons.
While humidity is a primary driver of precipitation, the intended deployment region for actuators is sparsely populated open ocean, where atmospheric moisture is continuously replenished from the sea surface and cannot be reduced on intervention timescales.
The proposed framework itself is variable-agnostic and naturally supports humidity perturbations when feasibility constraints permit.

\myparagraph{Baseline.}
We compare our method against the adversarial attack AOWF~\cite{Imgrund2025Adversarial}.
As precipitation-reduction intervention with data-driven diffusion forecasting models is a new task, no directly competing method exists; AOWF is, to our knowledge, the only public perturbation-based adversarial method for weather forecasting, making it our closest reproducible baseline.
AOWF optimizes $(\boldsymbol{\delta}^{t}, \boldsymbol{\delta}^{t+1})$ for both inputs $(\mathbf{X}^{t}, \hat{\mathbf{X}}^{t+1})$, but we restrict it to $\boldsymbol{\delta}^{t+1}$ to match \cref{sec:problem_setting}.
The attack budget is set to $\varepsilon = 0.07$ in normalized units, following the original implementation.
We enforce this constraint via a projection operator that limits the spatial standard deviation of each perturbed atmospheric variable.
The optimization is performed for 50 iterative update steps.
For both our method and AOWF, we use $n=2$ steps in the approximate inference, and fix the target time step to $T=2$ in~\cref{eq:problem_formulation}.

\myparagraph{Metrics.} 
Within the target region, we evaluate intervention performance using the reduction ratio and success rate.
The reduction ratio measures the relative decrease in 12-hour total precipitation compared to the standard forecast.
The success rate is the fraction of grid points that exceed the threshold $\tau$ in the standard forecast but fall below $\tau$ after intervention.
Outside the target region, we compute RMSE and MAE relative to the standard forecast to quantify unintended forecast changes.

\subsection{Qualitative evaluation}
\begin{figure}[t]
    \centering
    \includegraphics[width=\linewidth]{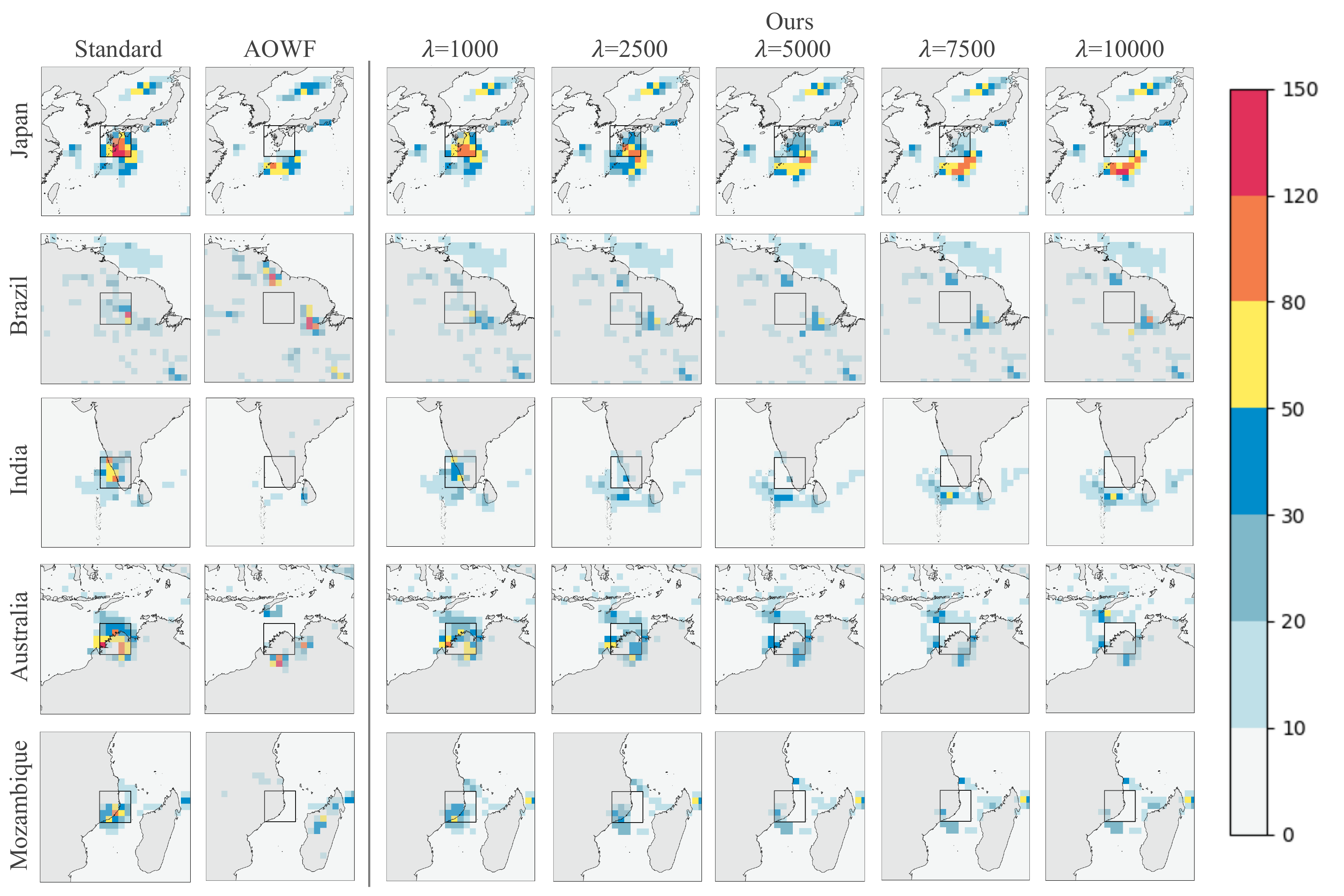}
    \caption{
        Qualitative evaluation of intervention effects.
        We compare the standard forecast, AOWF, and our method.
        The rectangle marks the target region.
        AOWF nearly suppresses precipitation within the target region.
        Our method reduces precipitation within the target region more gradually, increasing it in nearby regions.
        The colorbar reports 12-hour total precipitation in mm.
    }
    \label{fig:qualitative_evaluation}
\end{figure}

Figure~\ref{fig:qualitative_evaluation} shows qualitative comparisons of the standard forecast and forecasts after the interventions by AOWF and our method.
The standard forecast features extreme precipitation within the target regions (i.e., the boxed regions).
AOWF almost completely suppresses precipitation within the target region.
Our method reduces precipitation moderately, but a larger guidance scale $\lambda$ yields a more substantial reduction. 

Notably, the precipitation reduction from our method leads to an increase in precipitation in neighboring regions. 
Namely, our intervention diverts precipitation from the target region to neighboring regions, which is a reasonable consequence from a physical perspective. 
However, this is not the case for AOWF; the forecasted precipitation almost completely vanishes with no effect on neighboring regions (particularly in the India and Mozambique cases), which is unrealistic.

\subsection{Quantitative evaluation}
Table~\ref{tab:quantitative_evaluation} reports quantitative results averaged over the entire dataset, including precipitation reduction within the target region and the impact of the intervention on forecasts outside the target region.
AOWF achieves the largest precipitation reduction ratio and the highest success rate.
However, it also produces the largest RMSE and MAE outside the target region, indicating substantial forecast distortion.
Our method consistently attains lower RMSE and MAE than AOWF across all guidance scales $\lambda$.
These results indicate that our method better preserves forecasts outside the target region. 
Note that the quantitative metrics of the non-target region are computed over the entire globe, excluding the target region, whereas Figure~\ref{fig:qualitative_evaluation} focuses only on neighboring areas.

The guidance scale $\lambda$ controls the strength of the intervention.
At $\lambda = 1000$, reduction is limited while RMSE and MAE are minimized.
As $\lambda$ increases, both the reduction ratio and the success rate improve, reaching a success rate of $1.0$ at $\lambda = 7500$.
However, RMSE and MAE also increase accordingly.
These findings demonstrate a trade-off between suppression strength within the target region and forecast fidelity outside it. 

To summarize, the intervention by AOWF is powerful but apparently unnatural. 
Our method is relatively moderate in precipitation reduction, but it shows more physically plausible consequences and preserves forecast fidelity outside the target region. 
To provide a clearer contrast between AOWF and our method, we next conduct three sanity tests of the intervention.

\begin{table}[t]
    \centering
    \scriptsize
    \caption{
        Evaluation of target-region reduction and non-target degradation, averaged over the dataset.
        We report the reduction ratio and success rate (target), and RMSE and MAE (non-target).
        Our method reduces non-target degradation relative to AOWF.
        Increasing $\lambda$ yields a trade-off between control strength and non-target degradation.
    }
    \label{tab:quantitative_evaluation}
    \begin{tabular}{lcccc}
      \toprule
      \multicolumn{1}{l}{\textbf{Method}} &
      \multicolumn{2}{c}{\textbf{Target region}} &
      \multicolumn{2}{c}{\textbf{Non-target region}} \\
      \cmidrule(lr){2-3}\cmidrule(lr){4-5}
       & \textbf{Reduction ratio}$\uparrow$
       & \textbf{Success rate}$\uparrow$
       & \textbf{RMSE ($\times 10^{-4}$)$\downarrow$}
       & \textbf{MAE ($\times 10^{-6}$)$\downarrow$} \\
      \midrule
      AOWF~\cite{Imgrund2025Adversarial}
       & {\bfseries 0.9491}
       & {\bfseries 1.000}
       & 15.95
       & 435.0 \\
      Ours ($\lambda=1000$)
       & 0.3136
       & 0.7818
       & {\bfseries 0.9397}
       & {\bfseries 2.973} \\
      Ours ($\lambda=2500$)
       & 0.5414
       & 0.9420
       & 1.765
       & 4.945 \\
      Ours ($\lambda=5000$)
       & 0.7198
       & 0.9972
       & 2.596
       & 8.149 \\
      Ours ($\lambda=7500$)
       & 0.8025
       & {\bfseries 1.000}
       & 3.097
       & 8.446 \\
      Ours ($\lambda=10000$)
       & 0.8479
       & {\bfseries 1.000}
       & 3.446
       & 10.45 \\
      \bottomrule
    \end{tabular}
\end{table}
\section{Sanity Test 1: Profile of Interventions}\label{sec:structural_properties_of_the_interventions}
The first sanity test examines the intervention profile across atmospheric variables and pressure levels.
From a meteorological perspective, precipitation formation is primarily driven by dynamical and thermodynamical processes in the lower to mid-troposphere~\cite{Doswell1996FlashFloodForecasting,kato2018representative,Shibuya2021Dynamics}. 
Therefore, interventions focused on these layers are considered more physically plausible. 
While the upper troposphere can indirectly influence precipitation development through temperature profile and static stability, its contribution is relatively limited in mesoscale extreme precipitation events. 

\subsection{Setup}
\myparagraph{Metrics.}
We evaluate the magnitude and sparsity of the perturbation $\boldsymbol{\delta}^{t+1}$ at the surface and at each pressure level.
We measure magnitude using the spatial $\ell_2$ norm.
We define sparsity as the fraction of grid points $(p,q)$ where $|\delta^{t+1}_{p,q}| < 0.01$.
\begin{figure}[tb]
    \centering
    \includegraphics[width=\linewidth]{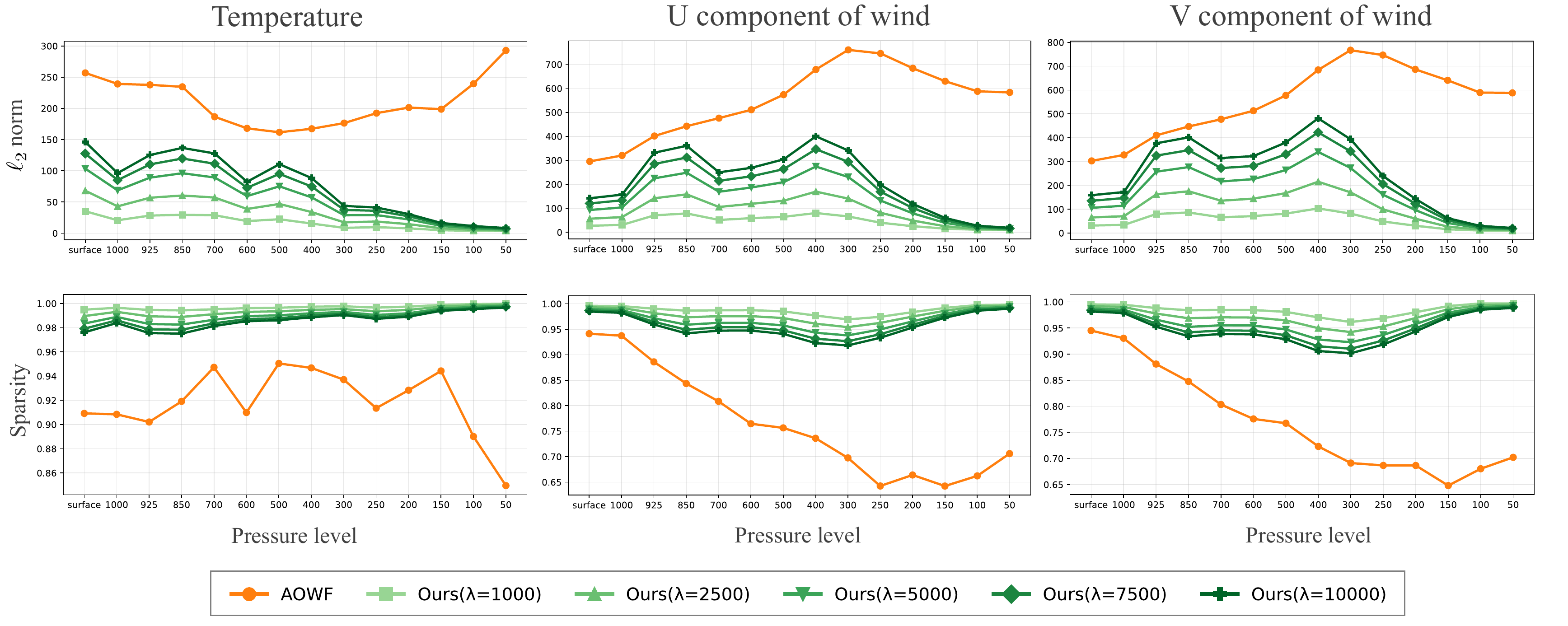}
    \caption{
        Perturbation profiles of $\boldsymbol{\delta}^{t+1}$, averaged over the dataset.
        Comparison of temperature and u and v components of wind.
        AOWF distributes perturbations across altitudes with the largest magnitudes appearing at upper levels for $T$ and wind, where mesoscale extreme precipitation has limited dependence.
        In contrast, our method concentrates perturbations at the low-to-mid troposphere where extreme precipitation primarily forms.
    }
    \label{fig:quantitative_intervention}
\end{figure}

\subsection{Structural properties of the perturbations}
Figure~\ref{fig:quantitative_intervention} shows the profiles of the intervention $\boldsymbol{\delta}^{t+1}$, averaged over the dataset.
For AOWF, the $\ell_2$ norm is non-uniform across pressure levels but reaches its largest magnitudes at upper levels for both temperature and wind.
This vertical distribution does not align with the dynamics of mesoscale extreme precipitation, which primarily originates from the lower to mid-troposphere.

In contrast, our method generates perturbations whose magnitude is concentrated at lower-to-mid pressure levels for both temperature and wind, while remaining smaller than AOWF's at all swept guidance scales $\lambda$.
Temperature perturbations concentrate at low levels, while wind perturbations peak at mid-levels.
These vertical patterns are smooth with respect to pressure levels.
This behavior is consistent with the understanding that low and mid-levels are more directly associated with precipitation formation.
Furthermore, our method produces sparser perturbations than AOWF across all guidance scales $\lambda$.
Here, sparsity is measured globally per pressure level and characterizes the low overall spread of the perturbation across the grid, not a pointwise localization at sub-grid scale.
Because the perturbations are produced by gradient guidance, their spatial profiles are smooth and do not contain the sharp spatial gradients that would be required to excite spurious gravity-wave-like artifacts.

The $\ell_2$ norm increases with $\lambda$, and perturbations become less sparse, indicating a stronger, more widespread intervention. 
Consistent with this, stronger interventions lead to greater precipitation reduction, as shown in Table~\ref{tab:quantitative_evaluation}.

Overall, our method yields locally and structurally organized perturbations that align with the atmospheric structure across variables and altitudes.

\section{Sanity Test 2: Latent-Space Deviation}\label{sec:latent_space_deviation_under_intervention}
The second sanity test examines the trajectories of GenCast's internal representations through the diffusion sampling steps. 
The GenCast model, a state-of-the-art data-driven weather forecasting model, is expected to approximate the atmospheric distribution well through its internal representations. 
Therefore, severe deviation from the trajectory without intervention implies unnatural changes in atmospheric dynamics.

\subsection{Setup}
\begin{figure}[tb]
    \centering
    \includegraphics[width=\linewidth]{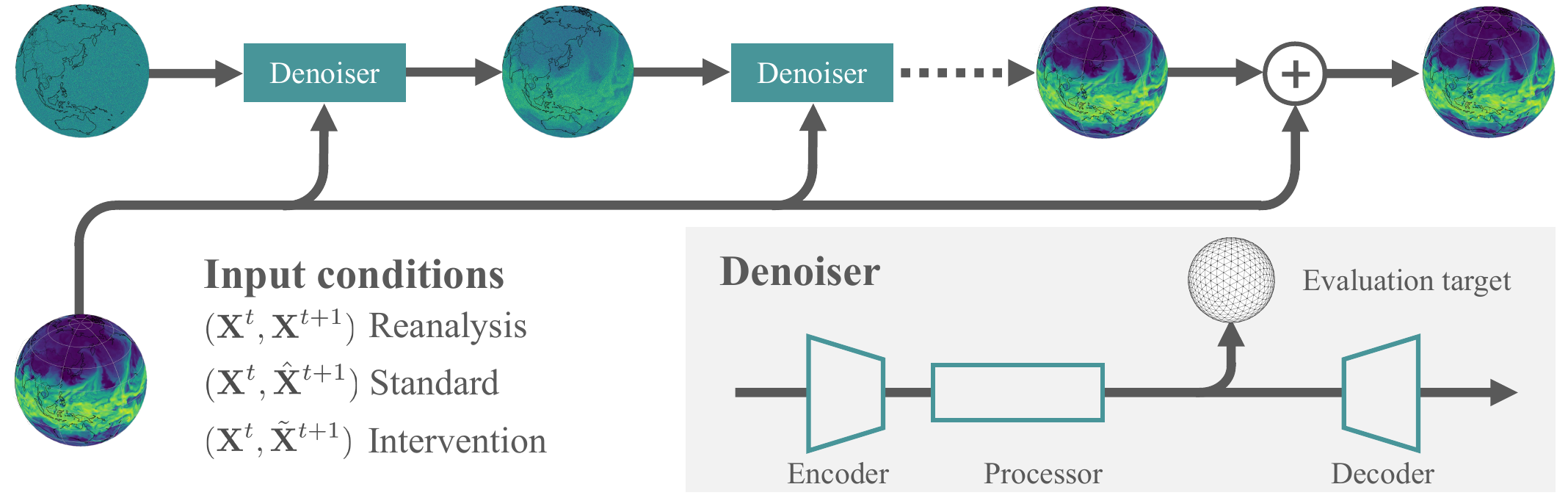}
    \caption{
        Illustration of the latent-space evaluation framework.
        Reanalysis, standard, and intervention inputs are propagated through diffusion sampling.
        At each step, the processor output is extracted as the latent representation.
        Analysis of these representations quantifies deviations in the latent space caused by the intervention.
    }
    \label{fig:proposed_evaluation_method}
\end{figure}

\myparagraph{Evaluation method.}
We extract latent representations from the denoiser at each sampling step, as illustrated in Figure~\ref{fig:proposed_evaluation_method}.
Specifically, we use the processor outputs of the GenCast denoiser as latent features.
We compare latent trajectories across three inputs: reanalysis inputs $(\mathbf{X}^t,\mathbf{X}^{t+1})$, the standard forecast $(\mathbf{X}^t,\hat{\mathbf{X}}^{t+1})$, and the intervention forecast $(\mathbf{X}^t,\tilde{\mathbf{X}}^{t+1})$.
We quantify deviations of the standard and intervention latents from the reanalysis latents at each sampling step.
Additional details are provided in the supplementary material.
\begin{figure}[tb]
    \centering
    \includegraphics[width=\linewidth]{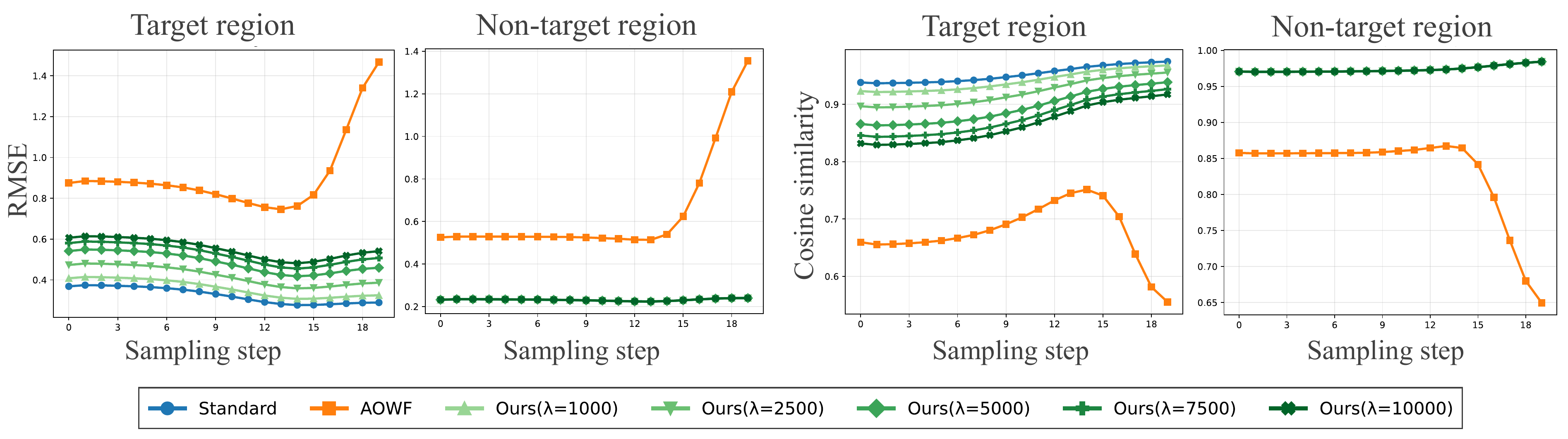}
    \caption{
        Evolution of latent-space RMSE and cosine similarity across sampling steps. 
        The standard forecast remains stable. AOWF exhibits sharp metric degradation in later steps. 
        Our method shows changes only within the target region depending on the guidance scale $\lambda$, while non-target regions remain similar to the standard forecast. 
    }
    \label{fig:latent_evaluation}
\end{figure}

\myparagraph{Metrics.}
We quantify the deviation of the standard and intervention latents from the reanalysis latents using root mean squared error (RMSE) and cosine similarity.
RMSE measures the magnitude of latent-space deviation, while cosine similarity evaluates the directional alignment.
We compute these metrics separately inside and outside the target region at each sampling step.

\subsection{Latent-space Dynamics under Intervention}
Figure~\ref{fig:latent_evaluation} shows latent-space RMSE and cosine similarity over sampling steps.
The standard forecast achieves the lowest RMSE and highest cosine similarity, remaining closest to the reanalysis latent representation.

AOWF exhibits large fluctuations within the target region.
Although it initially approaches the reanalysis latents around steps 13–14, both metrics worsen sharply afterward, indicating substantial deviation in later steps.
A similar pattern appears outside the target region, suggesting spillover effects.

Our method increases RMSE and decreases cosine similarity as $\lambda$ increases.
However, it consistently achieves lower RMSE and higher cosine similarity than AOWF.
RMSE stabilizes after step 14, and cosine similarity increases in later steps, indicating that the latent structure is preserved during denoising.

Outside the target region, both metrics match the standard forecast.
This indicates that the intervention is spatially localized while maintaining latent-space consistency.

\subsection{Visualizing the Latent-Space Trajectories}
\begin{figure}[tb]
    \centering
    \includegraphics[width=\linewidth]{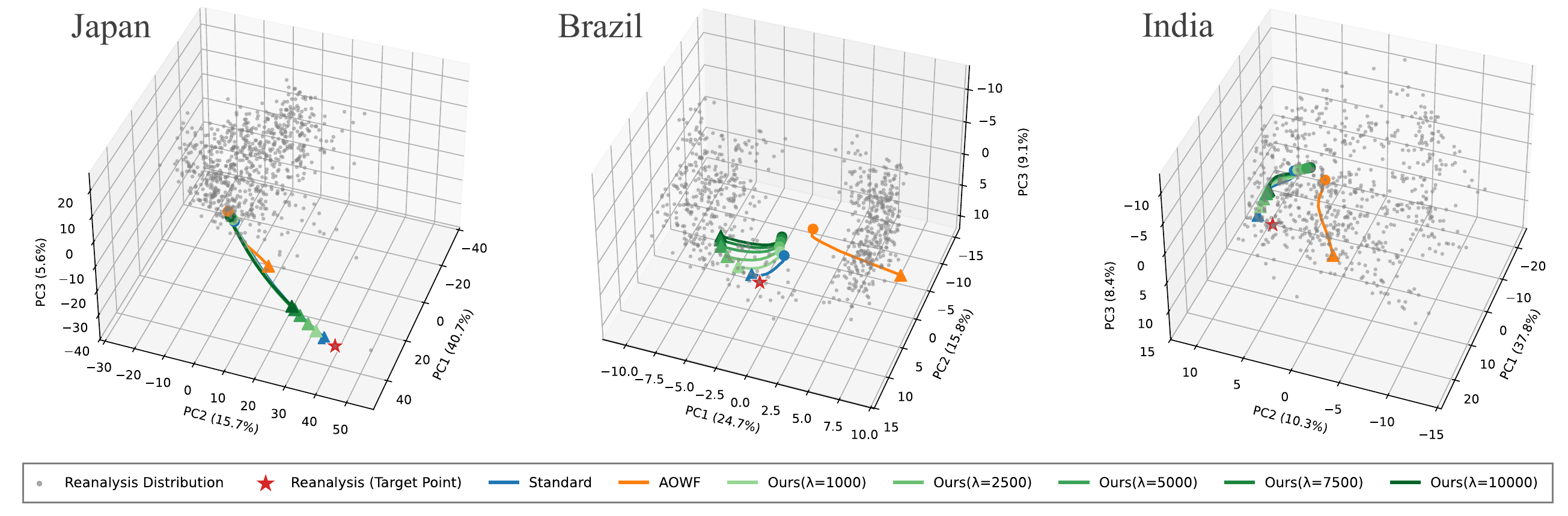}
    \caption{
        This evaluation visualizes the latent trajectory leading to the control effect shown in Figure~\ref{fig:qualitative_evaluation}.
        Latent representations are projected onto a 3D PCA space.
        Gray points show reanalysis latents in the target region, the reanalysis target point (red star) is for the selected sample, and arrows trace each method’s latent trajectory.
        AOWF trajectories tend to end in a different direction from, or farther from, the reanalysis target point than ours.
        Our method consistently steers latent trajectories toward the reanalysis target point.
    }
    \label{fig:latent_PCA_evaluation}
\end{figure}

Figure~\ref{fig:latent_PCA_evaluation} visualizes latent trajectories projected onto a three-dimensional principal component analysis (PCA) space.
Gray points denote the distribution of reanalysis latents within the target region across the dataset.
The reanalysis latent of the selected evaluation sample is denoted the reanalysis target point (red star).
Arrows indicate latent trajectories across sampling steps for each method.
Alignment is assessed by how closely the trajectory endpoints lie near the reanalysis target point.

The standard forecast follows a stable trajectory and ends near the reanalysis target point, indicating strong consistency with the reanalysis distribution.
In contrast, AOWF trajectories tend to end in a different direction from, or farther from, the reanalysis target point than ours; in the Brazil case, the AOWF endpoint lies in a different latent cluster.
This behavior is qualitatively consistent with the late-step degradation of latent-space RMSE and cosine similarity reported in Figure~\ref{fig:latent_evaluation}, which provides the quantitative counterpart of the trajectory visualization.

Our method consistently steers latent trajectories toward the reanalysis target point.
In Brazil, the trajectory remains within the cluster containing the reanalysis target point.
For smaller $\lambda$, trajectories approach the reanalysis target point more closely.
As $\lambda$ increases, the final distance gradually grows.
This suggests that stronger interventions induce larger deviations, yet remain closer to the reanalysis latents than AOWF.
\section{Sanity Test 3: Intervention Transferability}\label{sec:intervention_transferability}
The third sanity test examines the transferability of the intervention to GraphCast, another data-driven weather forecasting model with an architecture different from GenCast. 
If the intervention is not transferable, it indicates that the intervention is model-specific and thus not physically plausible.

\subsection{Setup}
We evaluate the transferability of $\boldsymbol{\delta}^{t+1}$ optimized for GenCast on GraphCast~\cite{Lam2023Learning}.
GraphCast is a graph neural network with an architecture and temporal resolution different from GenCast.
While GenCast produces 12-hour forecasts in a single step, GraphCast outputs 6-hour forecasts per step.
Accordingly, we add $\boldsymbol{\delta}^{t+1}$ to $\hat{\mathbf{X}}^{t+1}$ (i.e., $\tilde{\mathbf{X}}^{t+1}=\hat{\mathbf{X}}^{t+1}+\boldsymbol{\delta}^{t+1}$) and feed $(\mathbf{X}^{t}, \tilde{\mathbf{X}}^{t+1})$ into GraphCast for two sequential forecasting steps.
We then evaluate precipitation changes within the target region at each of the two 6-hour forecasting steps.

\subsection{Qualitative Evaluation}
\begin{figure}[tb]
    \centering
    \includegraphics[width=\linewidth]{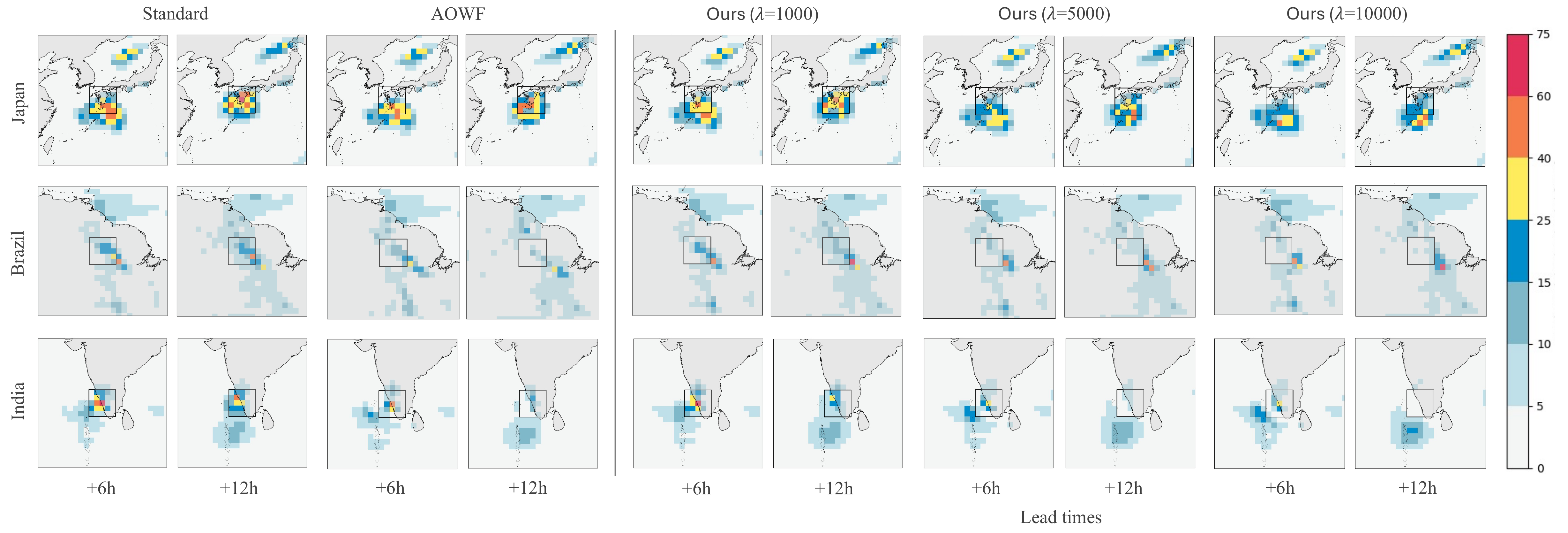}
    \caption{
        Qualitative evaluation of perturbation transferability to GraphCast.
        AOWF reduces precipitation in the India and Brazil cases but fails to achieve a sufficient reduction in the Japan case.
        In contrast, our method clearly reduces precipitation in India and Brazil and partially reduces it in Japan.
        The colorbar reports 6-hour total precipitation in mm, matching GraphCast's per-step output.
    }
    \label{fig:qualitative_transferability_control}
\end{figure}

Figure~\ref{fig:qualitative_transferability_control} shows the forecasts obtained by applying perturbations $\boldsymbol{\delta}^{t+1}$ optimized for GenCast to GraphCast.
In the standard forecast, GraphCast also produces extreme precipitation within the target region.
The transferability of AOWF is case-dependent.
It reduces precipitation in the India and Brazil cases but fails to achieve a sufficient reduction in the Japan case.

In contrast, our method exhibits better transferability than AOWF.
It reduces precipitation in India and Brazil and partially reduces it in Japan.
Moreover, precipitation is redistributed to surrounding regions, consistent with the qualitative evaluation in Figure~\ref{fig:qualitative_evaluation}.
The partial (rather than full) reduction in the Japan case reflects the intrinsic difficulty of cross-model transfer for typhoon-driven extreme precipitation, which is sensitive to the storm structure and is known to differ across data-driven forecasting models even without intervention.
These results suggest improved cross-model transferability compared to AOWF.
\section{Discussion and Future Work}
This work is the first comprehensive study of precipitation-reduction intervention using data-driven models, covering dataset collection, method design, and numerical evaluation.
As deep learning models can be readily fooled by adversarial perturbations, and since no ``ground-truth'' is available for the intervention and its consequences, it is crucial to carefully examine the physical plausibility of the generated interventions. 
To address this, we conducted three plausibility tests: (1) profiling the intervention, (2) latent-space deviation in GenCast, and (3) transferability to GraphCast.
From these perspectives, the experiments demonstrated that guiding the sampling trajectory of the GenCast model, rather than directly perturbing the atmospheric state as in adversarial attacks, leads to more physically plausible perturbations, as the generation process follows the atmospheric distribution approximated by the GenCast model.
The proposed framework is not tied to GenCast itself: it only requires a diffusion-based sampling process and therefore directly applies to future diffusion-based weather forecasting models.

\vspace{1mm}
For future work, the following are technical challenges for further enhancing the plausibility and feasibility of weather control.\footnote{This work focuses on the technical aspects of weather control. However, it is worth noting that the ethical implications and potential risks are also important areas of study, given their broad impact on society and the environment. } 

\myparagraph{NWP validation.}
A more rigorous test of transferability would be to feed the perturbed input into an NWP model and verify whether the intervention induces similar changes in the atmospheric state. 
However, this is not straightforward. Beyond differences in preprocessing between data-driven models and NWP systems, NWP models require dynamically balanced and physically consistent initial conditions~\cite{Bannister2017Review}.
Simply adding generated perturbations does not ensure growth under NWP dynamics.
Perturbations that are not dynamically balanced may rapidly decay during forecast integration~\cite{feng2020partition}.
Moreover, NWP forecasting itself involves careful configuration and tuning, even without additional perturbations.

\myparagraph{Flexibility of the intervention.}
Our method currently has limited control over the spatial distribution of the intervention. 
While the experiments demonstrated that the intervention is spatially localized and structured, greater flexibility in the intervention region is desirable. 
This is because feasible intervention regions may be restricted to sparsely populated areas such as oceans.

\section*{Acknowledgements}
This work was supported in part by JST Moonshot R\&D Grant Number JPMJMS2389 and JST BOOST Grant Number JPMJBY24C6.

\bibliographystyle{plainnat}
\bibliography{main}

\clearpage
\setcounter{section}{0}
\setcounter{figure}{0}
\setcounter{table}{0}

\appendix
\renewcommand{\thesection}{\Alph{section}}
\renewcommand{\thefigure}{\Alph{figure}}
\renewcommand{\thetable}{\Alph{table}}

\begin{center}
    {\LARGE \bf Appendix}
\end{center}
\vspace{2mm}

\section{Dataset Construction Details}\label{sec::dataset_construction_details}
This section provides additional details on the construction of the extreme precipitation event dataset used in our experiments.

\subsection{Extreme Event Extraction}
Following meteorological practice, we define extreme precipitation events as large deviations from climatology.
Specifically, for each calendar day, we compute the mean precipitation at each latitude–longitude grid point across all years in the WeatherBench2~\cite{Rasp2024WeatherBench} dataset and use this mean as the climatology.
The precipitation anomaly is then defined as the difference between the 12-hour total precipitation and this climatology.

Extreme precipitation events are identified based on precipitation anomalies. 
For each grid point, we compute the yearly maximum precipitation anomalies across the dataset and take the 99th percentile of these maxima.
The threshold is then defined as the spatial mean of these percentiles, yielding $\tau = 0.06668$, which provides a consistent criterion for event detection across regions.
Grid points where the anomaly exceeds $\tau$ are identified as extreme precipitation occurrences and are further restricted to land using the land-sea mask.
To prevent multiple detections of the same extreme precipitation event, events occurring within 48 hours and 300 km are treated as duplicates, as these thresholds reflect the typical spatial and temporal extent of extreme precipitation events.

\subsection{Predictability Filtering}
We restrict our evaluation to extreme precipitation events that are predictable by the forecasting model.
For each event, we run GenCast and evaluate whether the one-step forecast reproduces the extreme precipitation event.
If the maximum anomaly within the target region does not exceed the threshold $\tau$, the event is considered difficult for the model to forecast and is excluded from the evaluation dataset.
After applying this filtering procedure, the final dataset contains 219 extreme precipitation events, covering diverse regions such as Japan, Brazil, and India.

\section{Latent-space Analysis Details}\label{sec::latent_space_analysis_details}
This section provides additional details of the latent-space analysis used to evaluate the intervention.
We first describe the denoiser architecture of GenCast and then explain the evaluation procedure.

\subsection{GenCast's denoiser architecture}
In GenCast, the denoiser is implemented as a deep neural network consisting of an encoder, a processor, and a decoder.
The processor is a graph transformer that operates on a spherical mesh and models spatial interactions.
Geographic coordinates are explicitly incorporated in this architecture, with latitude and longitude information embedded in the nodes of the spherical mesh.
Consequently, the latent representations produced by the processor can be interpreted with respect to geographic locations on Earth.
For this analysis, we extract the processor outputs each time the denoiser is applied and use them as latent representations to analyze intervention effects.

\subsection{Evaluation Method}
We analyze how the model's latent representations evolve with and without intervention during forecasting toward time step $t+T$.
We compare latent representations across three inputs: the reanalysis input $(\mathbf{X}^t,\mathbf{X}^{t+1})$, the standard forecast $(\mathbf{X}^t,\hat{\mathbf{X}}^{t+1})$, and the intervention forecast $(\mathbf{X}^t,\tilde{\mathbf{X}}^{t+1})$.
For each input, we extract the processor outputs at each sampling step during forecasting from time $t+1$ to $t+T$ and use them as latent representations.
By analyzing these representations across sampling steps, we track how the influence of the intervention evolves during sampling.
We further separate the latent representations into those corresponding to nodes inside and outside the target region and analyze their representations separately.
This analysis evaluates whether the effects of the intervention remain confined to the target region or propagate to a broader area through the sampling process.

\section{Additional Results for Sanity Test 2}
\begin{figure}[t]
    \centering
    \includegraphics[width=\linewidth]{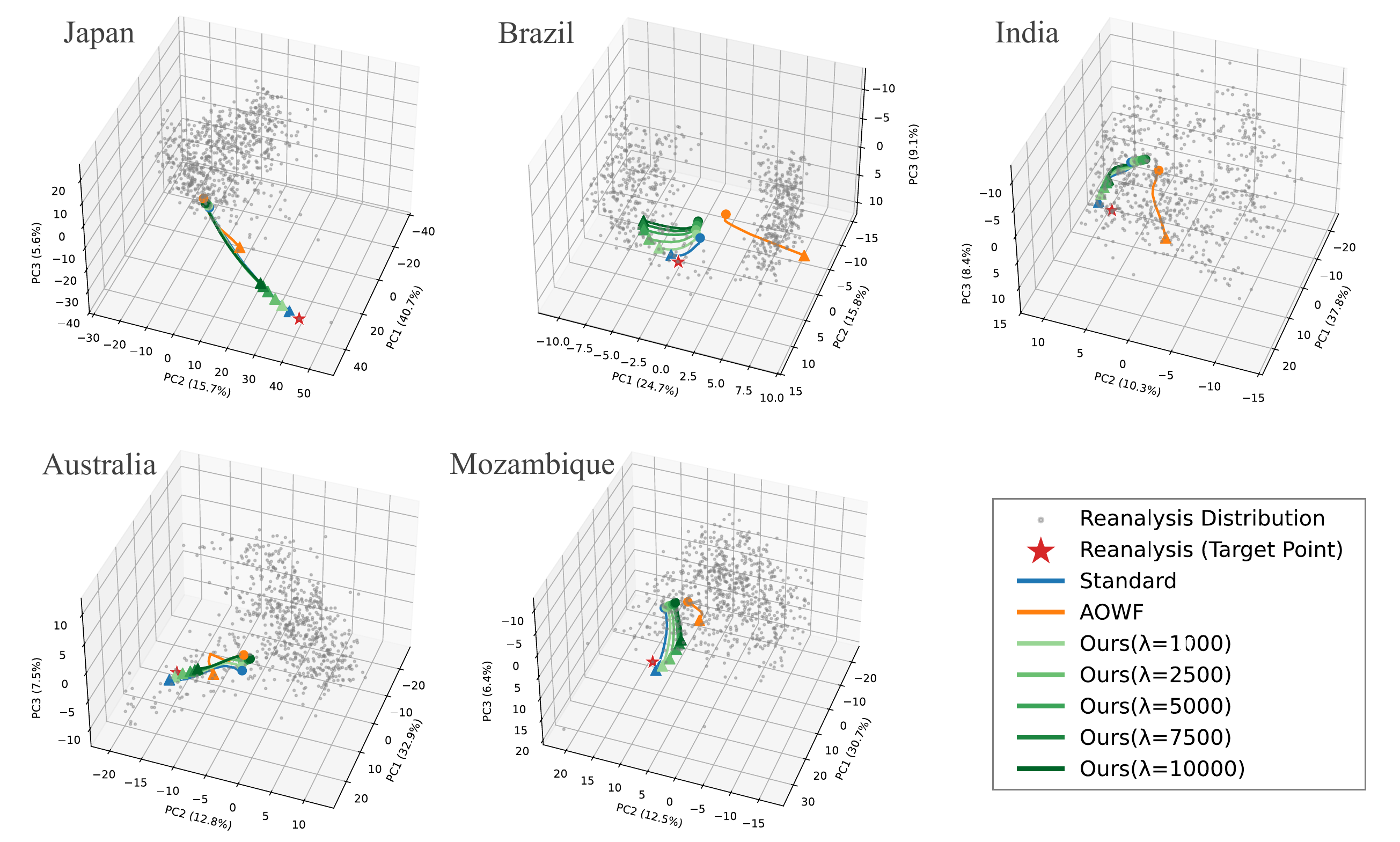}
    \caption{
        Complete visualization of the latent trajectories projected onto a three-dimensional PCA space.
        In addition to the cases shown in the main paper, we include additional cases from Australia and Mozambique.
        Gray points represent the distribution of reanalysis latents, and the red star denotes the reanalysis latent of the selected sample.
        While AOWF trajectories deviate from the red star, the proposed method consistently steers its latent trajectories toward the red star across different regions.
    }
    \label{fig:latent_PCA_evaluation_complete}
\end{figure}

Figure~\ref{fig:latent_PCA_evaluation_complete} presents the full visualization of the latent-space trajectory analysis for Sanity Test 2 in the main paper.
Due to space limitations, the main paper shows only a subset of cases.
Here, we provide results for additional regions to offer a more comprehensive view of the latent-space behavior.
Specifically, we include results for Australia and Mozambique in addition to Japan, Brazil, and India.
The additional cases show trends similar to those observed in the main paper.
AOWF tends to produce trajectories that move away from the red star, whereas the proposed method steers them toward the red star.
These results further suggest that the proposed method better preserves the latent trajectory of the forecast without intervention than AOWF.

\section{Additional Results for Sanity Test 3}
\subsection{Qualitative Evaluation}
\begin{figure}[t]
    \centering
    \includegraphics[width=\linewidth]{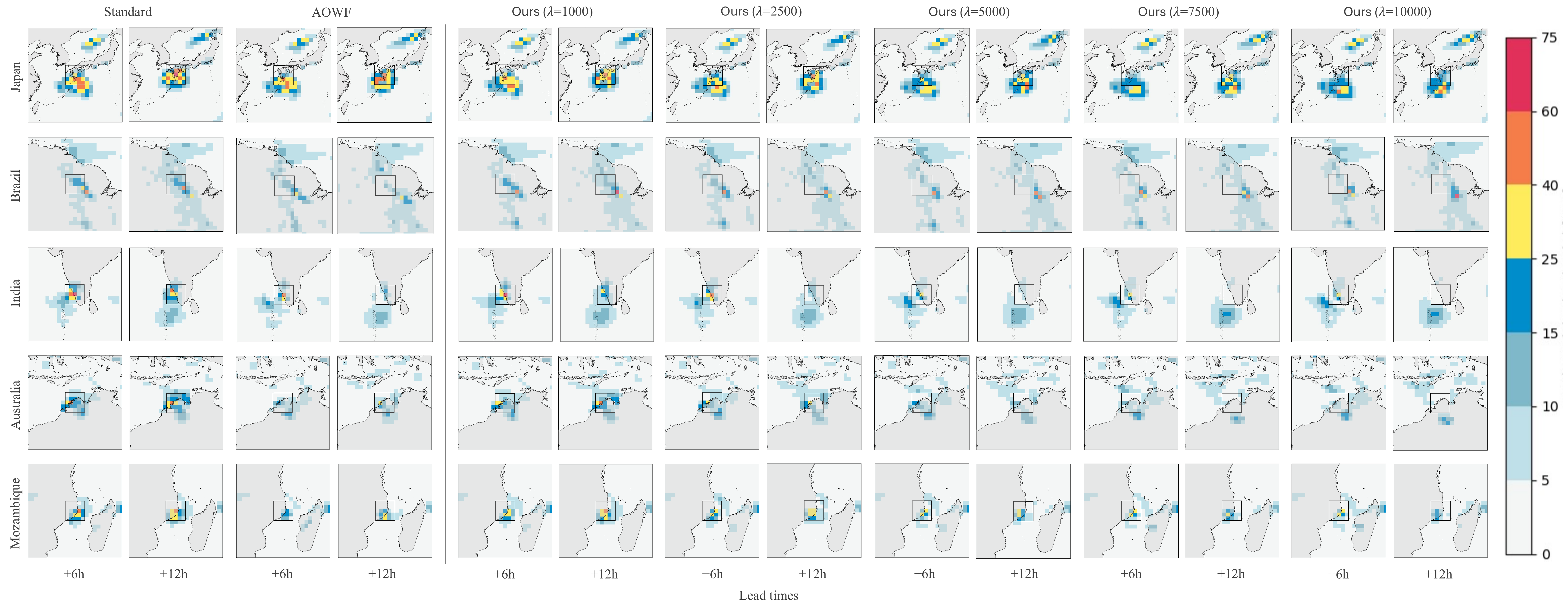}
    \caption{
        Complete qualitative results for the transferability evaluation.
        In addition to the representative cases shown in the main paper, we include additional regions and a wider range of guidance scales.
        The results further illustrate the improved cross-model transferability of the proposed method compared to AOWF.
    }
    \label{fig:qualitative_transferability_control_complete}
\end{figure}

Figure~\ref{fig:qualitative_transferability_control_complete} presents the complete qualitative results corresponding to the transferability evaluation in the main paper.
While the main paper shows only a subset of the cases due to space limitations, this figure provides additional regions and a finer range of guidance scales.
Specifically, we include results for Australia and Mozambique in addition to Japan, Brazil, and India, and evaluate multiple guidance scales.
These additional cases provide further evidence of the improved cross-model transferability of the proposed method compared to AOWF.

\subsection{Quantitative Evaluation}
Table~\ref{tab:quantitative_evaluation_graphcast} reports the quantitative results when the perturbations optimized for GenCast are transferred to GraphCast.
As GraphCast forecasts 6-hour total precipitation at each forecasting step, we run GraphCast for two consecutive steps and aggregate the two forecasts to obtain the 12-hour total precipitation.
The evaluation metrics are then computed using the same procedure as in the main paper.

Compared to the results in Table 1 of the main paper, AOWF shows decreases in both the reduction ratio and the success rate when transferred to GraphCast.
This indicates that adversarial perturbations exhibit limited cross-model transferability.
In contrast, the proposed method better preserves precipitation reduction performance than AOWF.
In particular, as the guidance scale $\lambda$ increases, the reduction ratio and success rate gradually improve.
At $\lambda = 10000$, our method achieves the highest reduction ratio and success rate among all methods.

These results suggest that the perturbations generated by the proposed method retain their precipitation reduction effect even when applied to a different forecasting model, demonstrating better cross-model transferability than adversarial perturbations.

\begin{table}[t]
    \centering
    \scriptsize
    \caption{
        Quantitative evaluation of perturbation transferability from GenCast to GraphCast.
        We report the reduction ratio and success rate for the target region, and RMSE and MAE outside the target region.
        The proposed method better preserves precipitation reduction performance than AOWF under cross-model transfer.
    }
    \label{tab:quantitative_evaluation_graphcast}
    \begin{tabular}{lcccc}
      \toprule
      \multicolumn{1}{l}{\textbf{Method}} &
      \multicolumn{2}{c}{\textbf{Target region}} &
      \multicolumn{2}{c}{\textbf{Non-target region}} \\
      \cmidrule(lr){2-3}\cmidrule(lr){4-5}
       & \textbf{Reduction ratio}$\uparrow$
       & \textbf{Success rate}$\uparrow$
       & \textbf{RMSE ($\times 10^{-4}$)$\downarrow$}
       & \textbf{MAE ($\times 10^{-6}$)$\downarrow$} \\
      \midrule
      AOWF~\cite{Imgrund2025Adversarial}
       & 0.4773
       & 0.8084
       & 1.862
       & 29.10\\
      Ours ($\lambda=1000$)
       & 0.1517
       & 0.4252
       & {\bfseries 0.8283}
       & {\bfseries 24.98} \\
      Ours ($\lambda=2500$)
       & 0.2970
       & 0.6729
       & 1.172
       &  26.02\\
      Ours ($\lambda=5000$)
       & 0.4338
       & 0.8084
       & 1.600
       &  27.24\\
      Ours ($\lambda=7500$)
       & 0.5116
       & 0.8738
       & 1.860
       & 28.07\\
      Ours ($\lambda=10000$)
       & {\bfseries 0.5614}
       & {\bfseries 0.8879}
       & 2.043
       & 28.66\\
      \bottomrule
    \end{tabular}
\end{table}

\section{Spatial Pattern Evaluation via Fractions Skill Score}\label{sec::fss}
The pointwise non-target RMSE/MAE used in the main paper can suffer from the so-called double-penalty effect, where a spatially shifted but otherwise plausible storm is penalized twice (a false alarm at the predicted location and a miss at the observed one).
To complement them with a more spatially tolerant view, we additionally evaluate the Fractions Skill Score (FSS)~\cite{Roberts2008Scale}, 
a standard spatial precipitation metric that compares binary occurrence fractions within a local neighborhood and is therefore lenient to small spatial displacements.

\subsection{Setup}
For each event, we binarize the standard forecast and the intervened forecast (AOWF or ours) using the extreme-precipitation threshold $\tau$ defined in~\cref{sec::dataset_construction_details}, then compute FSS over the non-target region using a $k^\circ \times k^\circ$ neighborhood with $k \in \{1, 3, 5\}$.
Higher FSS indicates better agreement of the surrounding precipitation pattern between the standard and intervened forecasts.

\subsection{Results}
\Cref{tab:quantitative_evaluation_fss} reports FSS at the three neighborhood scales.
Our method consistently achieves FSS\,$\geq 0.96$ across all guidance scales and all neighborhood sizes, while AOWF reaches only $0.47$--$0.58$.
The advantage holds even at the largest neighborhood ($5^\circ$), confirming that AOWF's degradation is not an artifact of small spatial shifts: AOWF disturbs the surrounding precipitation pattern, whereas our method preserves it.
This is consistent with the lower non-target RMSE/MAE reported in the main paper.

\begin{table}[t]
    \centering
    \scriptsize
    \caption{
        Fractions Skill Score (FSS) on the non-target region.
        FSS@$k^\circ$ denotes FSS computed with a $k^\circ \times k^\circ$ neighborhood, where binary occurrence is determined by the extreme-precipitation threshold $\tau$ defined in~\cref{sec::dataset_construction_details}.
        Higher is better.
    }
    \label{tab:quantitative_evaluation_fss}
    \begin{tabular}{lccc}
      \toprule
       & \textbf{FSS@$1^\circ$}$\uparrow$
       & \textbf{FSS@$3^\circ$}$\uparrow$
       & \textbf{FSS@$5^\circ$}$\uparrow$ \\
      \midrule
      AOWF
       & 0.4652
       & 0.5493
       & 0.5796 \\
      Ours ($\lambda=1000$)
       & {\bfseries 0.9901}
       & {\bfseries 0.9884}
       & {\bfseries 0.9829} \\
      Ours ($\lambda=2500$)
       & 0.9809
       & 0.9793
       & 0.9729 \\
      Ours ($\lambda=5000$)
       & 0.9669
       & 0.9662
       & 0.9625 \\
      Ours ($\lambda=7500$)
       & 0.9639
       & 0.9624
       & 0.9603 \\
      Ours ($\lambda=10000$)
       & 0.9590
       & 0.9581
       & 0.9571 \\
      \bottomrule
    \end{tabular}
\end{table}

\end{document}